\documentclass[11pt,a4paper]{article}
\usepackage[hyperref]{emnlp-ijcnlp-2019}
\usepackage{times}
\usepackage{latexsym}

\usepackage{url}

\usepackage{xcolor}
\usepackage{graphicx}
\usepackage{amssymb}
\usepackage{amsmath}

\aclfinalcopy

\title{Deep Label-Wise Attentive Temporal Convolutional Networks\\ Improve Medical Coding}

\author{Muhammed Yavuz Nuzumlal{\i} \\
  Yale University \\
  {\tt yavuz.nuzumlali@yale.edu} \\\And
  Alexander Fabbri \\
  Yale University \\
  {\tt alexander.fabbri@yale.edu} \\\AND
  Irene Li \\
  Yale University \\
  {\tt irene.li@yale.edu} \\\And
  Dragomir Radev\thanks{Deceased.} \\
  Yale University \\}

\date{}

\begin{document}
\maketitle

\begin{center}
\fbox{\begin{minipage}{0.93\columnwidth}
\small
\textbf{Note on the date of this work.} This work was carried out in 2019 at the LILY lab, Yale University, and the manuscript is posted here essentially as it stood at that time. The experiments and the baselines it compares against reflect the medical coding literature as of 2019. It is made available as a record of the work rather than as a current account of the field.
\end{minipage}}
\end{center}

\begin{abstract}
Medical coding is the task of assigning a set of diagnosis and procedure codes for a hospitalization using recorded notes. It requires aggregating information from different parts of the text and focus to different sections for each individual code, making it a very difficult problem even for professional human coders. We model the task as a multi-label text classification problem. To overcome the mentioned difficulties, we propose a deep neural model consisting of a multi-layer temporal convolution network (TCN) followed by label-wise attention. While multi-layer TCN helps extract a global document representation with the ability to learn relations over very long sequences, label-specific attention mechanism allows the model to focus on different aspects of the same document for each individual label. Our method achieves significantly better F-1 scores (9\% increase) compared to the previous state-of-the-art model, with a remarkable increase in recall score (28\% increase), which we believe is the more important metric for a clinical decision support setting.
%   We propose a novel deep multi-classifier neural network architecture for hierarchical label spaces. Each level of stacked blocks has a classifier layer that learns to discriminate the label set in the corresponding level of hierarchy, providing a guided learning procedure for the network. 
  %To handle very long sequences, we analyze the effects of a query-based content selection step using label descriptions as queries. 
%   For interpretability analysis, we extract the set of most important tokens for each label from the input sequence using attention weights for each classifier block. These are then compared with explanations provided by generic frameworks such as LIME~\cite{ribeiro2016should} and SHAP~\cite{lundberg2017unified}. 
%   We evaluate our approach on the medical coding problem, a multi-label classification problem that has a very large, imbalanced, hierarchical label space.
\end{abstract}

\section{Introduction}
Analyzing free-text Electronic Health Records (EHR) is a highly challenging task that has to be performed regularly by health care providers in clinical settings. Even though such records include extremely valuable information about the medical history of patients, their practical utilization is minimal because of the lack of standards, subjectivity among physicians and the labor intensive review process. Medical coding is a representative task which requires professional coders to manually and thoroughly review each individual recorded note for a hospitalization, with the aim of assigning a set of standardized International Classification of Diseases (ICD) codes corresponding to the procedures and diagnoses occurred during the hospitalization.

Apart from the challenges inherent in the complicated nature of clinical notes, such as different writing styles and lack of standardization, medical coding problem presents additional issues. First, the number of labels assigned to a specific note can be very large. Second, evidence for each individual label may exist at different locations within the note itself, requiring a person to aggregate global contextual information over distant sections of the notes rather than just using local contextual information. This is especially hard because of the length of the patient note records (the median number of words in a discharge summary in the MIMIC-III  dataset \cite{johnson2016mimic} is 1,388).

To tackle the problems described above, we propose a deep neural network architecture combining multi-layer temporal convolutional networks (TCN)~\cite{bai2018empirical} with a label-wise attention mechanism. Our model has ability to aggregate information over very long sequences thanks to convolution layers with exponentially increasing dilation factors; these dilations provide better awareness of the global context compared to the local context awareness of standard CNN models. Additionally, the label-wise attention mechanism enables the model to learn label specific document representations for patient notes by attending to different sections of the text sequence for each label. We call our method \textbf{L}abel-wise \textbf{A}ttentive \textbf{T}emporal \textbf{C}onvolutional \textbf{N}etworks (LATCN).

We evaluate our proposal on the freely accessible MIMIC-III dataset and compare with various related studies, including the SOTA models. Our model obtains significantly improved performance scores over the state-or-the-art models, especially in terms of recall, which we believe is a more important metric than precision for a clinical decision support system.

% which can be used for hierarchically structured label spaces to improve interpretability and accuracy. Specifically, we propose a deep attentive neural network architecture composed of small attentive neural network blocks stacked together, where each attentive network block learns to classify the set of labels from a particular level class hierarchy.

% We summarize our contributions as follows:

% \begin{itemize}
%     % \item We utilize recently proposed contextual pre-training models such as ELMo~\cite{peters2018deep} and BERT~\cite{devlin2018bert}, and analyze their effectiveness compared to static word embedding approaches such as word2vec~\cite{mikolov2013distributed} or GloVe~\cite{pennington2014glove}.
%     % \item 
%     % \item To overcome the difficulty of making direct inferences over very long language sequences such as patient notes, we propose a query-based content selection approach utilizing label descriptions as query keywords. \textcolor{red}{Or we can also have a weighted attention where prior weights come from ICD-9 descriptions as keywords?}
%     % \item We propose a novel network architecture, which we call Structured Transformer, for supervised learning on hierarchically structured label spaces, where we guide the network layers to learn label distributions via multiple classifier layers in the network.
% \end{itemize}

\section{Method}

\begin{figure*}[h!]
    \centering
    \includegraphics[width=0.8\textwidth]{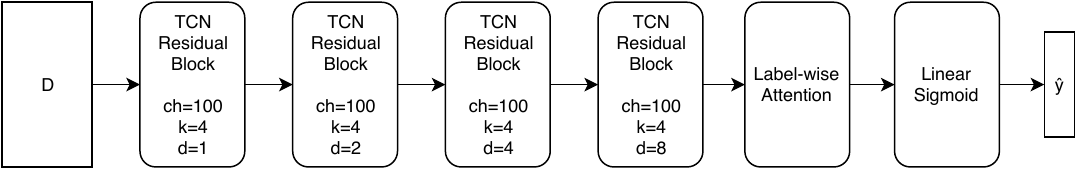}
    \caption{High level diagram of the proposed LATCN model.}
    \label{fig:tcn-attention}
\end{figure*}

We formulate the problem as a multi-label text classification task, where the aim is to assign a set of labels from the overall label space for each text document. The model consists of a multi-layer TCN followed by a label-wise attention mechanism, where we use word2vec \cite{mikolov2013distributed} embeddings pre-trained over the corpus of all discharge summaries as input.
The architecture is designed to extract a global context-aware representation for a given document by effectively passing information through the multi-layer TCN via exponentially dilated convolution filters. Using the TCN output as input, the  label-wise attention mechanism extracts per-label dense attentive representation vectors for the document, which are then used directly to predict class probabilities. A sketch of the architecture is shown in Figure~\ref{fig:tcn-attention}.

\paragraph{Temporal Convolutional Network}

We use TCN, as proposed in \cite{bai2018empirical}, for our base architecture to learn a matrix representation for a document. In our case, the TCN is composed of multiple residual blocks stacked together with a constant filter size $k$ and dilation factor of $d_\ell=2^\ell$ where $\ell$ is the index of the residual block. Each residual block includes 2 1-dimensional CNN layers with given parameters $k$ and $d$, where normalization, non-linearity, and regularization are provided by weight normalization~\cite{salimans2016weight}, rectified linear units (ReLU)~\cite{nair2010rectified}, and Dropout~\cite{srivastava2014dropout} functions respectively. This architecture provides a receptive field size of $2 \times (2^{\ell + 1} - 1) \times k$ for each hidden unit at layer $\ell$.

\paragraph{Label-wise Attention Mechanism}

After the TCN network transforms the input document matrix $D \in \mathbb{R}^{d_{e} \times N}$ ($d_e$ is the embedding vector dimension, $N$ is text sequence length) into the latent document matrix $H \in \mathbb{R}^{d_{ch} \times N}$ ($d_{ch}$ is the number of channels in convolutional layers), we apply a label-wise attention mechanism proposed by \citet{mullenbach2018explainable} to get individual document vectors for each label, which is the weighted summation of $H$ using attention weights per label. More formally, we learn an attention matrix parameter $U \in \mathbb{R}^{d_{ch} \times L}$ where $L$ is the number of unique labels. We calculate document vectors \(v_\ell\) for each label as follows:

\vspace{-4mm}
\begin{small}
\begin{align} 
A = & H^\top U, \\
\alpha_\ell = & \frac{exp(\alpha_\ell)}{\sum_{n = 1}^{N} exp(\alpha_{\ell n})}, &(softmax) \\
v_\ell = & \sum_{n = 1}^N \alpha_\ell h_n & (document\ vector)
\end{align} 
\end{small}

\vspace{-5mm}

\paragraph{Output Layer}
To calculate the label probabilities, we use one final linear output layer with a sigmoid activation function. We perform forward propagation of inputs in a different way than a classical fully connected layer such that each document vector \(v_\ell\) feeds only the output unit corresponding to label $\ell$. We can formulate this as follows:

\vspace{-6mm}
\begin{small}
\begin{align}
\small
\hat{y}_\ell = \sigma(w_\ell^\top v_\ell + b_\ell)
\end{align}
\end{small}

\vspace{-5mm}

\paragraph{Loss Function} We use binary cross-entropy loss, a standard method for multi-label classification, which is defined as:
\begin{equation}
\scriptsize
\begin{aligned}[b]
\mathcal{L}_{BCE}(X,y) = - \sum_{\ell=1}^{L}y_{\ell}log(\hat{y}_\ell) & + (1-y_\ell)log(1-\hat{y}_\ell)
\end{aligned}
\end{equation}

\section{Experiments}

\subsection{Dataset}
% \textbf{PubMed}\footnote{\url{https://www.ncbi.nlm.nih.gov/pubmed/}} indexes the MEDLINE database of references and abstracts on life sciences papers. PubMed plays a very important role in pre-training the models (such as language models in the BioBert model \cite{lee2019biobert}). 

% \textbf{PMC}\footnote{\url{https://www.ncbi.nlm.nih.gov/pmc/}} is a full-text archive of biomedical and life sciences literature at from NIH/NLM, with 5.3 million articles. Since we work on clinical notes, PubMed and PMC are only used to pre-train our language models. 

\textbf{MIMIC-III}\footnote{\url{https://mimic.physionet.org}}~\cite{johnson2016mimic} is a freely accessible dataset containing data from intensive care unit (ICU) hospitalizations over 10 years. For each hospitalization, the dataset contains a set of encounter notes recorded by health care providers, along with a set of assigned ICD-9 diagnosis and procedure codes and other structured information. Summary statistics can be found in Table ~\ref{tab:mimic-stats}.

\begin{table}[h!]
\small
\centering
\begin{tabular}{|c|c|}
\hline
\textbf{Hospitalizations} & 58,362 \\
\hline
\textbf{Total Notes} & 2,083,180 \\
\hline
\textbf{Discharge Summaries} & 59,652 \\
\hline
\textbf{Min document length} & 105 \\
\hline
\textbf{Median document length} & 1,388 \\
\hline
\textbf{Max document length} & 7,567 \\
\hline
\end{tabular}
\caption{\label{tab:mimic-stats} Summary statistics for MIMIC-III dataset.}
\end{table}
 
Following previous studies, we use only discharge summaries as input because they are the most comprehensive notes taken by a physician and serve as a summary of the entire hospitalization period. We also truncate texts to have a maximum length of 2,500 tokens. For the scope of this study, we focus on the 50 most common labels observed in MIMIC-III dataset, leaving the evaluation on the full label set as future work.

\subsection{Evaluation Metrics}

Although there are many previous studies on ICD code classification, no standard set of metrics exists. While AUC-ROC (Area Under The Curve - Receiver Operating Characteristic) and micro/macro averaged F1 are applied by most prior work, precision at $k$ (P@k) and recall at $k$ (R@k) are reported with inconsistent $k$ values, making results hard to compare and interpret. While P@k can be helpful to assess the correctness of the assigned values offered by a decision support system~\cite{mullenbach2018explainable}, R@k assesses how likely a system can prevent users from performing manual analysis of an entire document set to extract any possible missing values. In clinical settings, high recall is more valuable than high precision because the cost of searching for any missing code could be much higher than the cost of filtering among a set of codes.

To provide quantitative comparisons with previous work, we report R@5 and P@5 metrics, together with macro/micro averaged AUC-ROC, precision, recall, and F1 scores.

\subsection{Training}

We implement and train our model using the PyTorch\footnote{https://pytorch.org} library, with Adam~\cite{kingma2014adam} as the optimization method. For hyperparameters, we focused on tuning the number of TCN residual blocks $\ell \in \{1, 2, 4, 6\}$, number of channels $d_{ch} \in \{50, 100, 200, 300, 400, 500, 600\}$, and filter size $k \in \{4, 10\}$ using grid search. We selected best performing combination on development dataset which is $\ell=4$, $d_{ch}=100$, $k=4$. For the other hyperparameters, we followed the suggestions of the original TCN paper \cite{bai2018empirical}. Specifically, we use a dynamic learning rate with the \textit{reduce on plateau} policy from 0.002 towards 0.0001, and dropout rate of 0.4. We train for 200 epochs.

\vspace{-2mm}
\section{Results}

The overall performance metric scores for our model and previous studies are presented in Table~\ref{tab:mimic-50-results}. In general, our best performing method significantly outperforms the SOTA model DR-CAML for all metrics but macro/micro precision. F1 goes up 9\% on average.

We observe the most significant improvement on recall scores. Our model improves the result of DR-CAML approximately 28\%, which is a highly remarkable improvement. We hypothesize that the strong improvement on recall results from TCN's ability to learn better generalized models. We think that the very large receptive field of the proposed TCN model provides better global contextual information gains over the long text sequences compared to CNN models, which can only learn a very tight local contextual information of at most the size of the filter. The reduction in precision scores (approximately 7\%) also results from the same phenomenon; while a more generalized model can provide better coverage of the overall label space, a less generalized model can learn to predict a specific subset of labels very accurately, although having diminished performance over the remaining set of labels.

\paragraph{Discussion} In most clinical settings, we believe recall is a more important metric than precision, mainly because any automated system would only be useful as a decision support tool which will be validated by experts. This scenario will continue to be the same especially for NLP on clinical texts for the foreseeable future, given the performance gap between SOTA systems and human performance. Therefore, we suggest that automated systems should prioritize recall over precision.

\begin{table*}[h!]
\centering
\resizebox{2\columnwidth}{!}{\begin{tabular}{lcc|cc|cc|cc|cc}
\hline
 & \multicolumn{2}{c}{ROC-AUC} & \multicolumn{2}{c}{F-1} & \multicolumn{2}{c}{P} & \multicolumn{2}{c}{R} & & \\
Model & Macro & Micro & Macro & Micro & Macro & Micro & Macro & Micro & P@5 & R@5 \\
\hline
Logistic Regression & 0.829 & 0.864 & 0.477 & 0.533 & 0.546 & - & - & - & - & - \\
CNN & 0.876 & 0.907 & 0.576 & 0.625 & 0.620 & - & - & - & - & - \\
Bi-GRU & 0.828 & 0.868 & 0.484 & 0.549 & 0.591 & - & - & - & - & - \\
CAML~\citeauthor{mullenbach2018explainable} & 0.877 & 0.910 & 0.535 & 0.614 & 0.604 & \textbf{0.714} & 0.480 & 0.538 & 0.611 & 0.586 \\
% CAML~\citeauthor{mullenbach2018explainable} & 0.898 & 0.926 & 0.558 & 0.633 & 0.657 & \textbf{0.740} & 0.485 & 0.553 & 0.630 & 0.610 \\
LEAM~\citeauthor{wang2018joint} & 0.881 & 0.912 & 0.540 & 0.619 & 0.612 & - & - & - & - & - \\
DR-CAML~\citeauthor{mullenbach2018explainable} & 0.884 & 0.916 & 0.576 & 0.633 & \textbf{0.639} & 0.691 & 0.524 & 0.584 & 0.618 & 0.594 \\
LATCN (Our Model) & \textbf{0.908} & \textbf{0.931} & \textbf{0.631} & \textbf{0.681} & 0.595 & 0.655 & \textbf{0.672} & \textbf{0.710} &  \textbf{0.638} & \textbf{0.621} \\
% LATCN2 (Our Model) & \textbf{0.909} & \textbf{0.931} & \textbf{0.631} & \textbf{0.669} & 0.557 & 0.597 & \textbf{0.728} & \textbf{0.760} &  \textbf{0.642} & \textbf{0.622} \\
% LATCNFirst (Our Model) & \textbf{0.910} & \textbf{0.933} & \textbf{0.636} & \textbf{0.676} & 0.584 & 0.623 & \textbf{0.698} & \textbf{0.739} &  \textbf{0.643} & \textbf{0.624} \\
\hline
\end{tabular}}
\caption{\label{tab:mimic-50-results} MIMIC-III performance results on set of most common 50 labels. }
\end{table*}

% \begin{table*}[h!]
% \centering
% \small
% \resizebox{2\columnwidth}{!}{\begin{tabular}{lcc|cc|cccc|ccc}
% \hline
%  & \multicolumn{2}{c}{ROC-AUC} & \multicolumn{2}{c}{F-1} & \multicolumn{4}{c}{P@k} & \multicolumn{3}{c}{R@k}\\
% Model & Macro & Micro & Macro & Micro & 1 & 8 & 10 & 15 & 1 & 8 & 10 \\
% \hline
% Logistic Regression & 0.561 & 0.937 & 0.011 & 0.272 & - &  0.542 & - & 0.411 & - & - & - \\
% CNN & 0.806 & 0.969 & 0.042 & 0.419 & - & 0.581 & - & 0.443 & - & - & - \\
% Bi-GRU & 0.822 & 0.971 & 0.038 & 0.417 & - & 0.585 & - & 0.445 & - & - & - \\
% Match-CNN Ens. & 0.760 & 0.965 & 0.043 & 0.468 & \textbf{0.488} & 0.570 & - & - & 0.449 & \textbf{0.422} & - \\
% DR-CAML~\citeauthor{mullenbach2018explainable} & \textbf{0.897} & 0.985 & 0.086 & 0.529 & - & 0.690 & - & 0.548 & - & - & - \\
% CAML~\citeauthor{mullenbach2018explainable} & 0.895 & \textbf{0.986} & 0.088 & \textbf{0.539} & - & \textbf{0.709} & - & \textbf{0.561} & - & - & - \\
% ZAGCNN & - & - & 0.038 & - & - & - & 0.587 & - & - & - & 0.439 \\
% LATCN (Our Model) & 0.873 & 0.982 & \textbf{0.089} & 0.456 & 0.364 & 0.642 & - & 0.511 & \textbf{0.610} & 0.334 & - \\
% \hline
% \end{tabular}}
% \caption{\label{tab:mimic-full-results} MIMIC-III performance results on full label set (8922 labels). }
% \end{table*}

\vspace{-2mm}
\section{Related Work}

\subsection{Multi-Label Text Classification}
The task of extreme multi-label text classification extends multi-label classification to cases of massive label spaces, presenting problems of sparsity and scalability \cite{Liu17extreme, you18extreme}. \citet{Liu17extreme} present a family of CNN models to tackle a Wikipedia document classification dataset, while \citet{you18extreme} propose a bi-directional long short-term memory network and a multi-label attention mechanism to capture long distance dependencies and word-label importance.

\subsection{ICD Coding}
Medical coding presents similar challenges to extreme multi-label classification because of the enormous label space ($\sim$13K for ICD-9, $\sim$68K for ICD-10) and label imbalance in existing datasets.
\par
Approaches to ICD classification have largely focused on convolutional architectures. \citet{rios2018emr} combine matching networks with a CNN to improve frequent and infrequent label cases. 
\citet{mullenbach2018explainable}, meanwhile, use an attentional CNN (CAML). It extracts a local context-aware representation of the document, and attention weights learned per label select the most important segments of the document for a particular label. They also combine label embeddings pre-trained using code descriptions (DR-CAML) to improve performance. They match SOTA and use attention weights to offer interpretable predictions. \citet{rios2018few} extend the above approach using a graph-based CNN architecture to leverage structured relations among ICD-9 codes and learn hierarchical structure.
% The model selects a set of support documents by ranking via Euclidean distance between input vectors. It then feeds in the support documents together with the training example.
% They get improved results compared to various baseline models.

% Interpretability analysis using an expert physician shows that most activated 4-gram for each document and asking to an expert, where proposed method gets highest informativeness score.

% To make predictions, they combine global label vectors learned by GCNN with per-label document representations learned by ACNN. Fine-grained analysis on few-shot and zero-shot label performance shows significant improvements over previous studies.
Other approaches have similarly attempted to make use of the characteristics of the input and label spaces. \citet{wang2018joint} learn a joint word-label embedding space to build a compatibility metric between label-word pairs, while \cite{baumel2018multi} suggests a hierarchical attention model with Bi-GRU network blocks over sentences and documents. \citet{xu2018multimodal}, on the other hand, build an ensemble model to use unstructured (patient notes and ICD guidelines), semi-structured (ICD code descriptions), and structured (lab results, prescriptions) data. 

% They use a slightly different attention mechanism on top of a document matrix, and attention weights are learned using compatibility values between label and words as input, rather than directly using word vectors.
% This paper is not directly comparable with other recent studies because it only uses hand-selected 32 ICD codes as training examples, which significantly simplifies the actual problem.

\subsection{Temporal Convolutional Networks}
TCNs have outperformed RNNs on standard sequential benchmark tasks \cite{bai2018empirical}. Recently, this model was adapted to several classification tasks \cite{schwenk17deep, jiang18attention, abreu19hier}. \citet{schwenk17deep} use stacked temporal convolutional blocks to learn very deep networks for text classification.  \citet{abreu19hier} and \citet{jiang18attention} incorporate TCNs with attention for sentiment analysis and few-shot classification, respectively.

\vspace{-2mm}

\section{Conclusion}
We addressed the problem of assigning diagnosis and procedure codes for a hospitalization from encounter notes as a multi-label classification problem. We proposed a neural network architecture that incorporates temporal convolutional network blocks with label-wise attention to model longer text sequences. Our model outperforms the previous SOTA on a standard medical coding dataset and greatly improves recall scores compared to previous methods, a key characteristic in real-world medical settings. In future work we will experiment with this model on the full label set, in addition to investigating contextual embeddings and leveraging the hierarchical structure of ICD-9 codes.

\bibliographystyle{acl_natbib}
\bibliography{references}

\end{document}